\pdfoutput=1
\documentclass[11pt]{article}

\usepackage[final]{acl}

\usepackage{times}
\usepackage{latexsym}

\usepackage[T1]{fontenc}
\usepackage[utf8]{inputenc}

\usepackage{microtype}

\usepackage{sourcecodepro}

\usepackage{url}
\usepackage{booktabs}
\usepackage{graphicx}
\usepackage{amsmath}
\usepackage{tabularx}
\usepackage{ragged2e}
\usepackage{enumitem}
\usepackage{siunitx} % 只加载一次
\usepackage{listings}

\hypersetup{
    colorlinks=true,
    allcolors=blue, % 您可以根据喜好更改颜色，例如 allcolors=black
    pdfstartview={FitH},
}

\title{eSapiens: A Platform for Secure and Auditable Retrieval-Augmented Generation}

\author{Isaac Shi \and Zeyuan Li \and Fan Liu \and Wenli Wang \\
        \bf{Lewei He \and Yang Yang \and Tianyu Shi} \\
  eSapiens Team \\
        \url{https://www.esapiens.ai/}
        }

\begin{document}
\maketitle
\begin{abstract}
We present \textbf{eSapiens}, which is an AI-as-a-Service (AIaaS) platform engineered around a business-oriented trifecta: proprietary data, operational workflows, and any major agnostic Large Language Model (LLM).
eSapiens gives businesses full control over their AI assets, keeping everything in-house for AI knowledge retention and data security. eSapiens AI Agents (Sapiens) empower your team by providing valuable insights and automating repetitive tasks, enabling them to focus on high-impact work and drive better business outcomes.
The system integrates structured document ingestion, hybrid vector retrieval, and no-code orchestration via LangChain, and supports top LLMs including OpenAI, Claude, Gemini, and DeepSeek. A key component is the \textbf{THOR Agent}, which handles structured SQL-style queries and generates actionable insights over enterprise databases.

To evaluate the system, we conduct two experiments:\\
(1) A \textit{retrieval benchmark} on legal corpora reveals that chunk size of 512 tokens yields the highest retrieval precision (Top-3 accuracy: 91.3\%).\\
(2) A \textit{generation quality test} using TRACe metrics across five LLMs shows that eSapiens delivers more context-consistent outputs with up to \textbf{23\% improvement} in factual alignment.

These results demonstrate eSapiens’ effectiveness in enabling trustworthy, auditable AI workflows for high-stakes domains like legal and finance.\\
More details and demos are available at: \url{https://www.esapiens.ai/}.
\end{abstract}

\section{Executive Summary}

eSapiens is an enterprise‑grade AI “last‑mile delivery” platform that removes the gulf between cutting‑edge Large Language Models~(LLMs) and real‑world business operations. It combines a unified data fabric that ingests PDFs, Office documents and SQL tables,the DEREK engine that grounds every answer in verifiable context, and a visual no‑code workflow builder that lets non‑technical staff orchestrate multi‑step actions by simply asking questions in natural language. Under the hood, LangChain‑enhanced reasoning plans tool calls, executes SQL, invokes external APIs and appends source citations, while automatic vector indexing keeps content in sync without bespoke ETL pipelines. Security is enterprise‑first: end‑to‑end AES‑256 encryption, fine‑grained RBAC, VPC or on‑prem deployment options and SOC~2 Type II–aligned audit logging meet data‑residency and regulatory mandates, making the platform suitable for finance, insurance and life‑science workloads. A marketplace of more than 100 reusable prompts and 65 Quick Actions covers customer service, sales, finance, marketing and investment analysis, allowing pilot teams to launch new use‑cases in minutes rather than months and slashing development costs by up to 60 percent. Early adopters report that monthly financial reporting time fell from two hours to twelve minutes, automatic ticket categorisation accuracy rose by 40~percent and lead‑to‑deal velocity improved by double digits, demonstrating both rapid time‑to‑value and sustainable operational impact. With the global generative‑AI software market forecast to soar from USD 20 billion in 2024 to more than USD 130 billion by 2030 (CAGR $\approx$ 37 \%), enterprises urgently need a production‑grade bridge from model output to audited business action; eSapiens meets that need by delivering a secure, scalable and future‑proof foundation for data‑driven decision‑making, process automation and continuous AI innovation.
\begin{table}[!t]
\caption{Target Users and Core Value Proposition}
\small
\renewcommand{\arraystretch}{1.15}      % 行距
\setlength{\tabcolsep}{4pt}             % 列左右间距
\begin{tabularx}{\linewidth}{
    >{\RaggedRight\arraybackslash}p{0.26\linewidth}  % 约 26%
    >{\RaggedRight\arraybackslash}p{0.34\linewidth}  % 约 34%
    >{\RaggedRight\arraybackslash}X                 % 剩余 40%
}
\toprule
\textbf{User Persona} & \textbf{Current Pain} & \textbf{Value Delivered} \\\midrule
Data / Platform Engineers &
Heterogeneous ETL and DEREK pipelines; high run‑ops cost &
One‑click connectors plus automated vectorisation reduce integration and maintenance effort. \\[2pt]

Business Ops \& Service Teams &
Depend on SQL or scripts for reporting and ticket triage &
Natural‑language queries and no‑code workflows provide self‑service analytics in minutes. \\[2pt]

Security \& Compliance Officers &
Public LLM interfaces lack permissioning and audit trails &
VPC/on‑prem deployment, RBAC and citation traceability meet audit and data‑residency mandates. \\
\bottomrule
\end{tabularx}
\label{tab:persona}
\end{table}

\section{Market Background \& Challenges}
\subsection{Market Background}

Global spending on generative‑AI software and services has grown from less than USD 5 billion in 2020 to about USD 20 billion in 2024 and is projected to exceed USD 130 billion by 2030, reflecting a compound annual growth rate of $\approx$ 37\,\%.  Three strategic forces drive this trajectory.  \emph{Cost‑to‑serve compression}: retrieval‑augmented language models are able to automate 30–60\,\% of effort spent on document review, compliance reporting and first‑line customer support across banking, insurance and professional‑services sectors.  \emph{Real‑time decision loops}: executives increasingly demand instant synthesis of fragmented data streams—financial ledgers, risk dashboards, customer sentiment—turning AI from back‑office analytics into mission‑critical orchestration.  \emph{Customer‑experience differentiation}: buyers in a post‑pandemic digital economy expect personalised, context‑aware interactions, and brands deploying generative assistants routinely achieve double‑digit gains in satisfaction metrics.

Economic fundamentals accelerate adoption.  Token inference costs have fallen almost 40\,\% year‑over‑year due to transformer optimisation and commodity‑GPU availability.  Enterprise‑ready vector databases—Pinecone, Milvus, Elastic V7—have standardised high‑throughput similarity search, while orchestration frameworks such as LangChain and Semantic Kernel provide off‑the‑shelf tool calling and agent memory.  In parallel, regulatory activity is intensifying: the EU AI Act mandates risk classification and provenance tracking for high‑impact systems, updated GDPR guidance tightens data‑residency and audit‑logging, and sectoral rules (HIPAA, FFIEC, PCI‑DSS) extend existing security controls to generative workloads.  These shifts move enterprise attention from raw model quality to \emph{operational trust}: secure data access, citation‑level explainability and end‑to‑end governance.

The competitive landscape echoes this realignment.  Model‑as‑a‑Service providers deliver best‑in‑class reasoning but leave data integration to clients.  Pure‑play retrieval vendors solve storage and search complexity yet require external pipeline orchestration.  Vertical SaaS players embed narrow AI features—coding assistants, contract review, patient triage—but struggle to generalise.  Analysts highlight a persistent “middle‑layer gap”: platforms able to bridge private data, foundation models and downstream execution while meeting governance demands.  The gap is most acute in finance, healthcare and logistics, where document volumes are high, margins tight and compliance non‑negotiable.

\subsection{Core Challenges}

\begin{enumerate}[leftmargin=*,label=\arabic*.]
  \item \textbf{Data fragmentation and semantic gaps}\newline
        ERP, CRM, ticketing and document‑management systems expose no shared semantic layer or permission model, depriving language models of up‑to‑date context and forcing users into disruptive context switching.
  \item \textbf{High integration and run‑ops cost}\newline
        Bespoke vector indexes, DEREK pipelines and monitoring stacks typically require 3–6 months of multi‑disciplinary engineering effort, and ongoing maintenance erodes early return on investment.
  \item \textbf{Limited trust and explainability}\newline
        Hallucination risk and missing citation links undermine user confidence; output cannot be relied upon for finance, legal or life‑science decision workflows without verifiable provenance.
  \item \textbf{Talent and organisational gaps}\newline
        Business teams lack SQL and prompt‑engineering skills, while data teams lack domain context, causing projects to depend on a handful of AI champions and stalling scale‑out.
\end{enumerate}

By supplying a unified data fabric, a citation‑aware DEREK engine and a workflow designer within a zero‑trust security envelope, eSapiens directly addresses the middle‑layer gap, converting the four persistent obstacles above into a consolidated value proposition for large‑enterprise and mid‑market customers.

\section{Related Work}

The emergence of Retrieval-Augmented Generation (RAG) has significantly improved the factuality and traceability of large language model (LLM) outputs in knowledge-intensive domains \citep{lewis2020retrieval, izacard2020distilling}. Early frameworks like RAG \citep{lewis2020retrieval} and FiD \citep{izacard2020distilling} focus on tightly coupling retrievers with generators to improve response accuracy, but they are primarily research prototypes with limited support for enterprise deployment concerns such as access control, versioning, or compliance.

In the product space, LangChain \citep{langchain2022} and LlamaIndex \citep{llamaindex2022} have popularized modular toolchains for building LLM-powered applications. These systems provide developer-centric orchestration over retrieval, prompt formatting, and tool calling, yet require significant engineering effort to harden for production use. Similarly, Gorilla \citep{patil2023gorilla} and Toolformer \citep{schick2023toolformer} showcase autonomous agent behaviors and tool use but leave concerns like document-level traceability, permission management, and domain adaptation to the end user.

Commercial verticals have also produced domain-specific solutions, such as ChatLaw \citep{gao2023chatlaw} for legal QA and Lawyer-LLM \citep{fan2023lawyerllm} for Chinese regulatory domains. While impressive in scope, these systems are often single-purpose and difficult to generalize beyond their training scope.

\textbf{eSapiens} differentiates itself as a full-stack platform that addresses these gaps directly. It offers hybrid vector retrieval based on Elasticsearch, chunk-aware document indexing, and integrated prompt orchestration via LangChain, all encapsulated in a no-code UI layer and secured through role-based access control and tenant isolation. Compared to open frameworks and vertical tools, eSapiens emphasizes auditable, modular, and reusable AI infrastructure that reduces integration cost, accelerates time-to-value, and meets the security and governance standards required in finance, healthcare, and legal services.

\section{Product Overview}
\subsection{Definition}

eSapiens is an enterprise middle‑layer platform that securely links Large Language Models (LLMs) with proprietary data and day‑to‑day workflows.  The system unifies data ingestion, citation‑aware Retrieval‑Augmented Generation (RAG) and a drag‑and‑drop workflow builder, enabling business users to retrieve insights, automate multi‑step actions and audit every response from a single conversational interface.

\subsection{Key Differentiators}
\textbf{Customization}. eSapiens ships more than one hundred reusable prompt templates and sixty-five Quick Actions. Domain teams can fine‑tune terminology or JSON‑schema parameters without writing code.

\noindent
\textbf{Integration}. The platform provides one-click connectors for PDF, Office, SQL and mainstream SaaS sources, then performs automatic vectorisation and permission sync. End‑to‑end AES‑256 encryption, VPC deployment and role‑based access control enforce a zero‑trust posture.

\noindent
\textbf{Usability}. A familiar search‑bar + chat interface combined with a visual workflow designer lets business users launch use‑cases in under two weeks. A real‑time dashboard shows execution paths, source citations and token spend to simplify monitoring and optimisation.

\subsection{Comparison with Existing Platforms}

Current solutions fall into three categories.  \emph{Model‑as‑a‑Service} vendors provide state‑of‑the‑art reasoning yet leave data integration and governance to customers.  \emph{Pure‑play retrieval} products address storage and search complexity but require external orchestration and user‑interface layers.  \emph{Vertical SaaS} tools embed narrow AI features—coding assistants, contract review, clinical triage—though their tightly scoped designs hinder cross‑domain expansion.  eSapiens fills the resulting middle‑layer gap by combining secure data access, citation‑aware retrieval, visual orchestration and zero‑trust governance in a single extensible package, making it suitable for both large enterprises that need standardised AI infrastructure and mid‑market firms that cannot afford bespoke integration projects.

\section{Core Functional Modules}
\subsection{Sapiens Creation}

This module converts business intent into production‑ready AI agents (“Sapiens”) through three tightly integrated capabilities that emphasise reuse, speed and domain fit:

\begin{itemize}[itemsep=4pt]
  \item \textbf{Role‑based modular agent design} – An agent is assembled from independent role blocks that encapsulate a system prompt, allowed tool calls and guard rules.  Blocks can be combined like Lego pieces, enabling one finance analyst role to share its tool stack with multiple reporting bots while still inheriting global audit policies.  Version tags on each block provide Git‑style diff and rollback, so upgrades never break existing workflows.

  \item \textbf{No‑code customisation and deployment} – A three‑step wizard captures metadata, data sources and trigger logic; users finish the process by clicking \emph{Publish}.  Behind the scenes, the platform provisions vector indexes, signs API keys and rolls the agent out to SaaS, VPC or on‑prem Kubernetes clusters with blue–green safety.  Built‑in CI hooks export performance metrics (latency, cost, citation coverage) to popular observability tools.

  \item \textbf{Domain‑specific persona templates} – More than twenty pre‑configured personas—customer support, financial analyst, legal reviewer, medical scribe and R\&D researcher—ship with tuned prompts, industry lexicons and miniature eval sets to ensure factual accuracy from day one.  Templates can be cloned and extended with custom entities or forbidden topics, allowing a bank, for example, to layer its internal compliance rules on top of the generic “finance” persona without touching core code.  Continuous evaluation dashboards highlight drift and suggest prompt refinements as regulations or business terminology evolve.
\end{itemize}

\subsection{Prompt Management System}

\begin{itemize}[itemsep=4pt]
  \item \textbf{Prompt Template Library} – The platform provides a multi‑level prompt library that serves as a central repository for reusable instructions. Templates are grouped into three visibility scopes—public gallery, team library and personal drafts—and tagged by industry, role and task. Each entry contains the prompt text, a short abstract, sample output and basic metadata such as owner and last‑modified date. Users can search or filter by keywords, clone a template into their own workspace and edit it through a rich‑text editor that supports Markdown and JSON examples. 

  \item \textbf{Quick Action Prompts} – Quick Actions are lightweight, single‑purpose prompt snippets aimed at high‑frequency tasks—for example writing release notes, extracting contract highlights or generating go‑to‑market checklists. Each Quick Action exposes only a few parameters through a short form and can be inserted directly into the chat panel, workflow builder or Sapiens setup wizard. At run time the system injects the user‑supplied values and returns a ready‑to‑use answer within seconds, sparing frontline staff from composing lengthy prompts. All executions inherit the platform’s permission model and audit trail, so usage statistics and generated content are logged automatically for later review.
\end{itemize}
\subsection{Data Connectors \& Knowledge Management}

\textbf{Unified Data Connector Layer.}
More than twenty out‑of‑the‑box connectors—Outlook, SharePoint, OneDrive, Google Docs, Amazon S3, HubSpot, Slack, X/Twitter APIs and others—ingest content with a single OAuth or service‑account authorisation.  Each connector normalises files into a common document schema that records MIME type, timestamp, author and ACL metadata.  Webhook or change‑log triggers capture deltas every one to five minutes, so the knowledge base stays “near‑real‑time” with source systems.

\textbf{Knowledge Repository and Version Governance.}
A central dashboard displays file origin, size, vector‑token count, linked Sapiens agents and last access dates.  The repository applies immutable versioning: the first upload creates v1; any edit, re‑chunk or re‑index produces v2, v3, and so on.  Users can diff or roll back versions in one click, providing full auditability and protection against accidental loss.

\textbf{Semantic Index and Intelligent Retrieval.}
Text is chunked at paragraph level and written to a hybrid vector index that also stores structured fields; optional multimodal embeddings cover images and video.  Every query passes through pre‑filtering by ACL and tenant boundary, then post‑filtering by guard rules (regex, PII detection).  Results return the original file path, the matched passage and an extractive summary, so users can verify answers at source.

\begin{itemize}[leftmargin=*,itemsep=3pt]
  \item \textbf{Broad format coverage} – PDF, Word, Excel, CSV, e‑mails, chat logs, cloud blobs and SQL rows are supported without custom ETL.
  \item \textbf{Minute‑level synchronisation} – Delta polling or webhooks keep indices fresh, giving Sapiens access to the latest context.
  \item \textbf{Version traceability} – Immutable history, diff view and one‑click rollback answer the question “who changed what, and when”.
\end{itemize}

With this module, organisations gain an extensible ingest pipeline and a governed, searchable knowledge base, providing timely, secure and audit‑ready context for all downstream Sapiens agents and automated workflows.

\subsection{Security \& Access Control}

eSapiens implements a comprehensive, multi-layered security architecture to safeguard data privacy, particularly in scenarios involving proprietary documents, internal knowledge bases, and AI training data. The following components collectively ensure compliance, traceability, and robust defense against unauthorized access.

\textbf{Secure Transmission and Encryption.} All document uploads are protected via HTTPS and stored using Amazon S3 with AES-256 encryption, both in transit and at rest. These measures ensure that even in the event of a breach, unauthorized entities cannot access the content.

\textbf{Access Control and Tenant Isolation.} Fine-grained access control is enforced to restrict document operations—upload, modification, deletion—to authorized users only. The system supports multi-level permissions, including private bots and organization-scoped bots. Strong tenant isolation ensures that no cross-access or data leakage occurs in multi-tenant deployments.

\textbf{Knowledge Base Security.} Uploaded documents are immutably versioned and indexed under tenant-specific access policies. eSapiens continuously monitors usage patterns using New Relic, and all user actions are recorded in audit logs to support security investigations and compliance reviews. Periodic data backups ensure rapid recovery in case of loss or corruption.

\textbf{Vector Database and Embedding Security.} The Elastic Cloud vector database used for document indexing and retrieval runs on a certified AWS infrastructure. It features both physical and logical isolation to prevent unauthorized read, copy, modification, or deletion during data transmission and storage.

\textbf{LLM and Prompt-Level Safeguards.} AI models deployed within eSapiens are selected based on their security performance. Prompt data is securely stored and encrypted, and all access events are logged. The platform conducts regular model-level security reviews and applies sanitization to mitigate the risk of prompt injection or data leakage.

\textbf{System-Wide Defense.} Platform-wide protections include CAPTCHA-based registration checks, intrusion detection systems, regular security patching, and strict internal access controls. Employees undergo periodic security training. Redundancy and failover mechanisms are in place to guarantee high availability. Additionally, all API interactions include nonce values to defend against replay attacks.

These security practices establish eSapiens as a trustworthy and regulation-ready solution for enterprise-grade AI deployments in sensitive domains such as law, finance, and healthcare.

The platform supports enterprise‑grade governance by combining tenant isolation, role‑based access control (RBAC) and real‑time observability.

\begin{itemize}[leftmargin=*,itemsep=3pt]
  \item \textbf{Multi‑user permission model} – Four default roles (Owner, Admin, Editor, Viewer) cover most workflows, while custom roles allow fine‑grained scopes such as “read vector index only” or “execute but not edit workflow”.
  \item \textbf{Team‑based collaboration} – Workspaces can be partitioned by department or project; assets such as Sapiens agents, prompts and data connectors inherit team ACLs automatically, simplifying cross‑functional sharing.
  \item \textbf{Admin dashboard \& usage analytics} – A central console tracks API calls, token spend, latency and model mix (GPT‑4o, Claude 3, Gemini 1.5 etc.).  Drill‑downs show which user or agent consumed which model, enabling cost allocation and anomaly detection.
\end{itemize}

\subsection{API \& Embed Integration}

Developers can extend or embed platform capabilities through a secure, versioned interface stack.

\begin{itemize}[leftmargin=*,itemsep=3pt]
  \item \textbf{RESTful API for external systems} – Endpoints support query, generate, vector‑write and workflow‑trigger actions; authentication uses scoped API keys and optional OAuth 2.0 service accounts.
  \item \textbf{Embeddable UI components} – A lightweight JavaScript widget renders the chat panel or a single Sapiens agent inside Web apps; iframe and React hooks are also available for deeper integration.
  \item \textbf{Third‑party connectors} – Pre‑built adapters for Slack, Microsoft Teams, Notion and Atlassian Confluence let users invoke Sapiens, receive citations and push workflow outputs without leaving their native tool.
\end{itemize}

\subsection{DEREK: Deep Extraction and Reasoning Engine for Knowledge}

This module delivers highly efficient and precise document question-answering services by integrating advanced AI technologies, vector search, and intelligent recall mechanisms, optimizing the document retrieval and content generation workflow.

\noindent\textbf{Document Preprocessing and Vectorization.}  
Supports multiple file formats including Doc, Txt, PDF, and EPUB through LangChain-standard loaders. Documents are segmented into chunks of 1,000 tokens with a 150-token overlap. Each chunk is transformed into a high-dimensional vector using OpenAI Embeddings and stored in Elasticsearch Cloud, establishing an efficient hybrid vector index.

\noindent\textbf{Intelligent Recall Workflow:}  
\begin{itemize}[leftmargin=*,itemsep=3pt]
    \item \textbf{Query Refinement:} User queries are rewritten by GPT-4o for more precise semantic alignment.
    \item \textbf{Hybrid Retrieval Strategy:} Combines keyword-based and semantic similarity searches, returning the top 50 most relevant document snippets efficiently via Elasticsearch.
\end{itemize}

\noindent\textbf{Answer Generation and Quality Assurance:}  
\begin{itemize}[leftmargin=*,itemsep=3pt]
    \item \textbf{Prompt Optimization:} Utilizes the CO-STAR prompt format to clearly specify context, objectives, style, tone, and audience requirements, enhancing interaction with language models.
    \item \textbf{Multi-Agent Validation:} Employs LangGraph for secondary verification of answers. Insufficient answers trigger either online searches or re-generation requests to ensure high quality.
\end{itemize}

\noindent\textbf{Performance Advantages and Evaluation.}  
As demonstrated by experimental results in Appendix~\ref{sec:appendix-a} (Retrieval Performance) and Appendix~\ref{sec:appendix-b} (Output Quality Evaluation), the eSapiens platform significantly outperforms traditional FAISS-based methods, specifically in:
\begin{itemize}[leftmargin=*,itemsep=3pt]
    \item \textbf{Context Relevance:} Substantially improved, providing answers better aligned with user intent (see Appendix~\ref{sec:appendix-b}).
    \item \textbf{Utilization:} Higher effectiveness, particularly notable in GPT-4o and Gemini 1.5 Pro, enhancing the extraction and synthesis of critical information (see Appendix~\ref{sec:appendix-b}).
    \item \textbf{Accuracy and Naturalness:} Achieves higher scores in fluency, coherence, and professional alignment, as evidenced by user evaluations compared to traditional approaches (see Appendix~\ref{sec:appendix-b}).
\end{itemize}

The DEREK module significantly improves the accuracy and practicality of question-answering in enterprise scenarios, particularly suited for regulatory interpretation, report generation, and summarization tasks involving complex content.

\subsection{THOR: Transformer Heuristics for On-Demand Retrieval}

The THOR module is designed for structured data question answering tasks. By deeply integrating natural language understanding with structured query generation, it enables the automatic transformation of user-input natural language questions into executable SQL statements, along with result interpretation and feedback. This module provides a unified query interface for all table-centric data sources on the platform, significantly lowering the barrier for non-technical users to access complex information.

The overall architecture of the THOR module adopts a decoupled design, consisting of an orchestration layer and an execution layer, which ensures scalability and maintainability. The orchestration layer is responsible for receiving natural language input, identifying user intent and task types, and routing the requests to the appropriate downstream components. The execution layer comprises several functional agents, each responsible for SQL generation, result interpretation, and exception handling with self-correction.

Specifically, the SQL generation agent combines semantic understanding with awareness of database schema to accurately generate SQL queries, supporting complex conditions, aggregations, and multi-table joins. The result interpretation agent performs structured analysis and natural language generation based on the query results, extracting key values and trends and delivering human-readable answers. In case of query failure or unsatisfactory results, the self-correction module is automatically triggered. It reconstructs the query intent and applies multi-round generation strategies to enhance robustness and fault tolerance.

The THOR module has been successfully applied in scenarios such as report analysis, regulatory auditing, metric extraction, and business inspection, greatly improving the efficiency of structured data interaction. Its core capabilities encompass not only SQL generation and interpretation, but also contextual awareness, fault-tolerant generation, and expressive result delivery, forming a closed-loop natural language interface for structured databases that supports intelligent, data-driven decision-making for non-technical users.

\section{Technology Stack}

\begin{itemize}[leftmargin=*,itemsep=2pt]
  \item \textbf{LLM backbone} – A model router based on langchain\_openai presents one API to several engines. Supported models: GPT-4o mini, GPT-4o, GPT-4.1 (OpenAI); Claude 3.7 Sonnet and Sonnet Extended (Anthropic); Gemini 1.5 Pro (Google); DeepSeek-R1 and V3. Rate limits and retries are normalised so the rest of the code remains model agnostic.
  \item \textbf{DEREK engine and vector store} – Retrieval-Augmented Generation uses Elasticsearch 8.x with dense\_vector and HNSW. Documents are chunked by RecursiveCharacterTextSplitter, embedded with OpenAIEmbeddings and stored together with ACL metadata. A hybrid search that mixes BM25 and vector similarity improves recall on long legal text.
  \item \textbf{Prompt orchestration} – A stateless FastAPI service runs LangChain chains for prompt templating, tool calls (SQLAlchemy, webhooks, Slack) and guard rails such as regex PII filters. Execution graphs are streamed to Kafka for later monitoring and replay.
\end{itemize}

\section{System Architecture}

\begin{figure*}[htbp]
    \centering
    \includegraphics[width=0.95\linewidth]{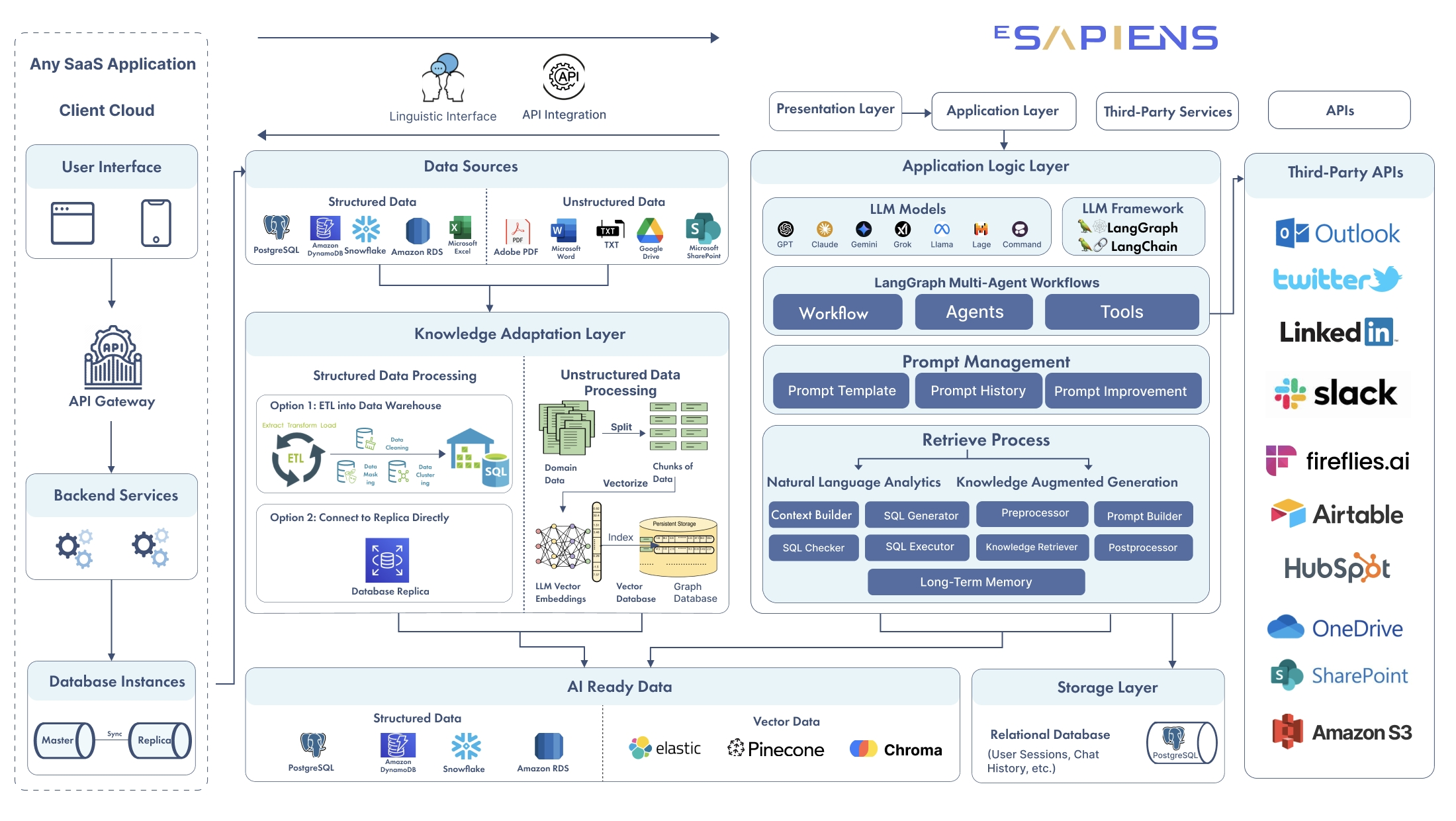}
    \caption{Overall Architecture of the eSapiens System}
    \label{fig:system-architecture2}
\end{figure*}

\subsection{Architecture Overview}
\label{sec:architecture-overview-1}

The \textit{eSapiens} system adopts a layered, modular architecture that enables seamless integration of natural language understanding, structured/unstructured data processing, and retrieval-augmented generation across enterprise applications. As shown in Figure~\ref{fig:system-architecture2}, the architecture spans from user interfaces and SaaS APIs on the frontend to backend services, databases, and multi-agent orchestration on the backend.

At the core lies the \textbf{Knowledge Adaptation Layer}, which bridges the gap between diverse data sources and downstream language model workflows. Structured data (e.g., PostgreSQL, Snowflake, RDS) is ingested either via ETL into centralized data warehouses or through direct connection to replicas, enabling SQL-based querying and analytics. Unstructured data (e.g., PDFs, Word documents, TXT files, SharePoint) undergoes domain-specific parsing and chunking, followed by embedding using LLM-based vectorization, and is stored in vector or graph databases.

The \textbf{Application Logic Layer} orchestrates multi-agent workflows via \textit{LangGraph} and \textit{LangChain} frameworks. It supports both structured queries (through SQL generation and execution) and unstructured question answering (via retrieval and LLM response generation). Prompt management modules track templates, histories, and refinement processes, while the retrieval pipeline includes context building, pre-/post-processing, and knowledge-grounded augmentation.

The architecture is \textit{model-agnostic}, supporting major foundation models including GPT, Claude, Gemini, and LLaMA, and integrates seamlessly with third-party services (e.g., Outlook, Slack, SharePoint) through API connectors.

Finally, all interactions are mediated through the \textbf{API Gateway} and stored in a central \textbf{storage layer}, which records user sessions, chat history, and retrieval logs in a relational database. This design ensures traceability, extensibility, and enterprise-grade deployment flexibility.

\begin{figure*}[htbp]
  \centering
  \includegraphics[width=0.9\textwidth]{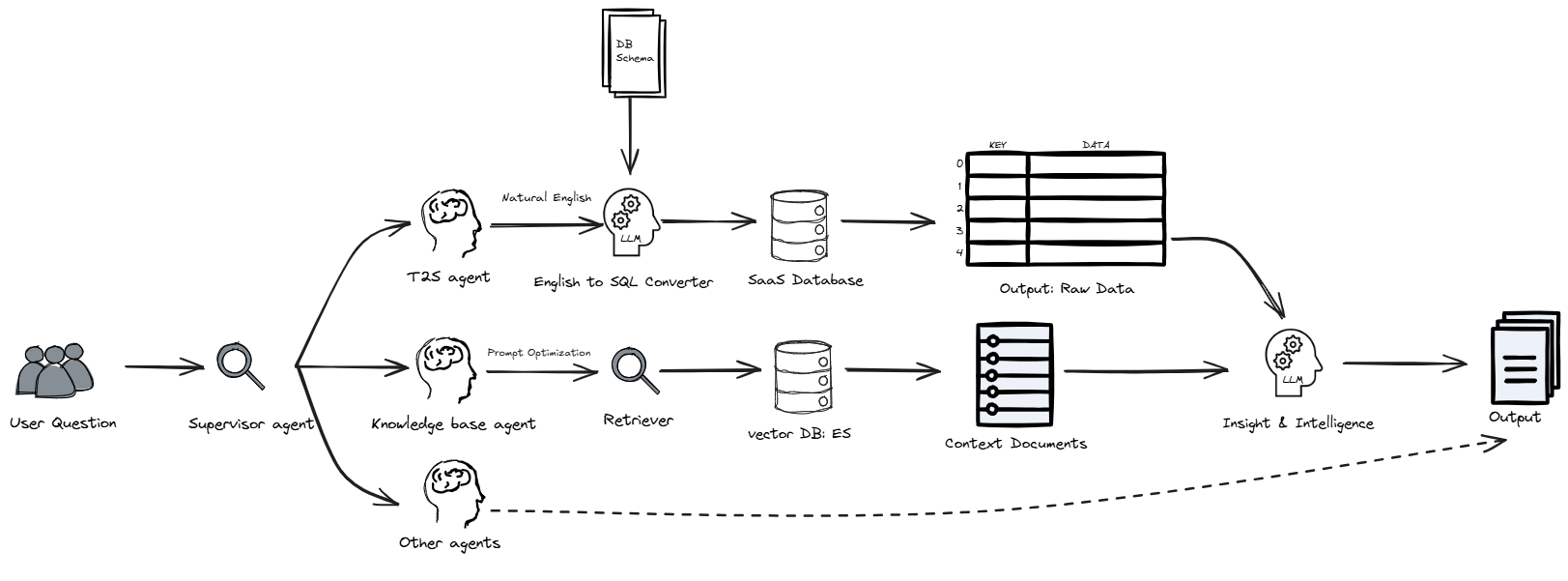}
  \caption{Overall Multi‐Agent System Architecture.}
  \label{fig:system-architecture}
\end{figure*}

\subsection{Overall Agent-Oriented System Architecture}
\label{sec:architecture-overview-2}

As shown in Figure~\ref{fig:system-architecture}, the overall system is built on a multi‐agent coordination framework.

\begin{itemize}
\item \textbf{Data Analyze Agent}: Handles fully‐structured, SQL‐based queries and insight generation.
\item \textbf{Knowledge Base Agent}: Manages retrieval‐augmented question answering over unstructured document stores.
\item \textbf{Other Agents}: Includes auxiliary modules such as web search and summarization agents.
\end{itemize}

Each path terminates in one or more LLM-powered functional agents that produce user-facing text output. All LLM interactions are orchestrated around our enhanced foundation model, \textbf{eSapiens-claude-3.7-extended}, a customized variant fine-tuned for long-context reasoning, knowledge grounding, and high-precision language generation. This model serves as the backbone for high-quality natural language outputs across different agent tasks.

\subsection{Knowledge base Agent}
\label{sec:DEREK}

This module targets retrieval‐augmented generation over unstructured documents.  It consists of:

\begin{itemize}
  \item \textbf{Supervisor Agent}: routes “document QA” queries to the Knowledge base Agent path.
  \item \textbf{Retriever}: performs a vector‐search (e.g.\ ElasticsearchStore) over indexed embeddings, returning the top‐\(k\) relevant snippets.
  \item \textbf{LLM Document Answer Agent}: consumes the retrieved context and the original question, then produces a coherent natural‐language answer.
\end{itemize}

\subsubsection{Operational Workflow}

\begin{itemize}
  \item \textbf{Query Dispatch}: Supervisor Agent identifies the query as “document QA” and forwards to DEREK Agent.
  \item \textbf{Vector Retrieval}: Retriever encodes the question, runs similarity search, and returns top‐\(k\) context documents.
  \item \textbf{Answer Generation}: Document Answer Agent prompts the LLM with question + context and generates the answer.
  \item \textbf{Response Delivery}: Final answer is returned to the user.
\end{itemize}

\subsection{Data Analyze Agent}

This module targets structured data question answering tasks and is built upon a multi-agent coordination architecture. The Data Analyze Agent enables dynamic reasoning over structured databases by leveraging transformer-based techniques, improving the system’s ability to convert natural language queries into executable SQL, interpret the results, and generate actionable insights. The architecture supports various business scenarios such as report analysis, regulatory auditing, and data-driven decision making, providing intelligent, context-aware interaction with enterprise data systems.

\subsubsection{System Architecture}

The Data Analyze Agent adopts a decoupled architecture that separates the orchestration layer from the execution layer. It consists of a task dispatcher and several specialized functional agents:

\begin{itemize}
\item \textbf{Supervisor Agent}: Responsible for receiving user queries, interpreting task types, and routing them to appropriate functional agents for downstream processing.
\item \textbf{Functional Agents}, including:
\begin{itemize}
\item \textbf{SQL Generation Agent}: Converts natural language questions into executable SQL queries based on database schema and semantic understanding.
\item \textbf{Result Interpretation Agent}: Analyzes the returned tabular data, extracts key values and trends, and generates semantic explanations in natural language.
\item \textbf{Self-correction Module}: Automatically triggered upon query failure or unsatisfactory results; re-generates SQL queries using scoring and retry mechanisms.
\end{itemize}
\end{itemize}

This modular design supports flexible extensions and integration with various data sources and structured data tasks.

\subsubsection{Operational Workflow}

The Data Analyze Agent follows a closed-loop workflow covering question analysis, SQL generation, execution, and insight synthesis. The process consists of the following steps:

\begin{itemize}
\item \textbf{Task Analysis and Routing}: Upon receiving a natural language question, the Supervisor Agent determines whether it requires structured query processing and dispatches it to the SQL Generation Agent.
\item \textbf{SQL Construction and Execution}: The SQL Generation Agent constructs an SQL statement by referencing schema structure and field semantics, then submits the query to the backend data warehouse.
\item \textbf{Self‑Correction \& Rating}: If the query returns an empty result, execution error, or a low‐quality rating, the Self‑Correction Module invokes the LLM to analyze and regenerate the SQL, then retries execution.
\item \textbf{Insight Generation}: Once a valid result is obtained, the Result Interpretation Agent analyzes the tabular output, extracts key values and trends, and converts them into user‐facing natural language insights.
\item \textbf{Answer Delivery}: The final answer is presented in a readable format, supporting either concise summarization or detailed explanatory output.
\end{itemize}

This workflow incorporates scoring and context-aware mechanisms to enhance answer quality, improve robustness, and provide interpretable outputs for complex structured queries.

\subsubsection{Self‑Correction \& Rating}
\label{sec:data-analyze-self-correction}

To enhance SQL execution success and improve result accuracy, the Data Analyze Agent includes a self‑correction and rating loop. When an empty result, execution error, or low‐quality rating is detected, the system automatically re‑generates and re‑executes SQL up to a fixed number of retries, guided by LLM feedback and scoring.

\section{Applications \& Use Cases}
\subsection{Use Case 1: Knowledge-Centric Question Answering over Enterprise Archives}

\textbf{Scenario Description.}  
Employees in large organizations often need to retrieve and comprehend policy, process, or technical details stored in archived documents. These documents are typically unstructured (PDFs, Word, scans) and dispersed across multiple platforms, making conventional keyword-based search ineffective for semantic understanding and content aggregation.

\textbf{Workflow and System Roles.}  
The user inputs a natural language question through the interface, such as ``What are the internal control requirements for audit processes?'' The system identifies the task as document-based QA and routes it to the \textit{Knowledge Base Agent}.

The \textit{Retriever} encodes the query and searches for semantically relevant text spans among pre-embedded document chunks, returning the top-\(k\) results. These snippets, along with the original query, are forwarded to the \textit{LLM Document Answer Agent}. The language model performs multi-passage reasoning to synthesize a coherent, grounded answer, which is returned to the user with optional source attributions.

\textbf{System Capabilities Demonstrated:}
\begin{itemize}
  \item Supports automatic segmentation, vectorization, and indexing of unstructured documents;
  \item Constructs dynamic prompts combining user intent, context, and retrieved snippets;
  \item Enables answer traceability by attaching source citations from retrieved content;
  \item Handles complex queries such as multi-turn questions or cross-topic document reasoning.
\end{itemize}

\vspace{1em}

\subsection{Use Case 2: Natural Language-Driven Structured Data Analysis}

\textbf{Scenario Description.}  
Business users often require fast access to key metrics and performance summaries from internal databases, but lack the SQL expertise or tooling familiarity to perform direct queries over complex relational schemas.

\textbf{Workflow and System Roles.}  
The user submits a query in natural language, such as ``What is the order exception rate by department this quarter?'' The system classifies this as a structured data task and delegates it to the \textit{Data Analyze Agent}.

The \textit{SQL Generation Agent} uses schema-aware intent interpretation to construct an executable SQL query. If execution fails (e.g., due to schema mismatch), the \textit{Self-Correction Module} is triggered to analyze the failure, regenerate the query using LLM feedback and retry.

Once results are retrieved, the \textit{Result Interpretation Agent} analyzes the tabular data, extracts relevant KPIs or patterns, and generates a natural language summary tailored to the user's request, optionally supporting detailed or high-level views.

\textbf{System Capabilities Demonstrated:}
\begin{itemize}
  \item Supports automatic mapping from natural language to SQL queries using schema-aware parsing;
  \item Provides semantic disambiguation and error correction for misaligned field names or query logic;
  \item Implements a feedback-driven loop for SQL regeneration and retry based on LLM evaluation;
  \item Enables result summarization with language generation over tabular outputs;
  \item Allows configurable granularity in generated answers, supporting concise overviews or detailed breakdowns.
\end{itemize}

% Bibliography
\bibliography{references}

\begin{thebibliography}{8}
\providecommand{\natexlab}[1]{#1}

\bibitem[{Chase(2022)}]{langchain2022}
Harrison Chase. 2022.
\newblock Langchain: Language models in chains.
\newblock \url{https://www.langchain.com/}.

\bibitem[{Fan et~al.(2023)}]{fan2023lawyerllm}
Weicheng Fan and 1 others. 2023.
\newblock Lawyer llm: An expert-level chinese legal large language model.
\newblock \emph{arXiv preprint arXiv:2310.10472}.

\bibitem[{Gao and et~al.(2023)}]{gao2023chatlaw}
Xiaofei Gao and et~al. 2023.
\newblock Chatlaw: Open-source legal large language model trained on chinese legal documents.
\newblock \emph{arXiv preprint arXiv:2305.14251}.

\bibitem[{Izacard and Grave(2021)}]{izacard2020distilling}
Gautier Izacard and Edouard Grave. 2021.
\newblock Distilling knowledge from reader to retriever for question answering.
\newblock In \emph{ICLR}.

\bibitem[{Lewis and et~al.(2020)}]{lewis2020retrieval}
Patrick Lewis and et~al. 2020.
\newblock Retrieval-augmented generation for knowledge-intensive nlp tasks.
\newblock In \emph{NeurIPS}.

\bibitem[{Liu(2022)}]{llamaindex2022}
Jerry Liu. 2022.
\newblock Llamaindex (gpt index).
\newblock \url{https://www.llamaindex.ai/}.

\bibitem[{Patil et~al.(2023)}]{patil2023gorilla}
Chinmay~H. Patil and 1 others. 2023.
\newblock Gorilla: Large language model connected with massive apis.
\newblock \emph{arXiv preprint arXiv:2305.15334}.

\bibitem[{Schick et~al.(2023)}]{schick2023toolformer}
Timo Schick and 1 others. 2023.
\newblock Toolformer: Language models can teach themselves to use tools.
\newblock \emph{arXiv preprint arXiv:2302.04761}.

\end{thebibliography}

\appendix
\section{Retrieval Performance on Long-form Legal QA}
\label{sec:appendix-a}

\subsection{Objectives and Scope}
This benchmark evaluates the eSapiens DEREK pipeline on long-form legal QA tasks, with a focus on four datasets. It aims to answer three core questions:

\begin{itemize}[leftmargin=*,itemsep=2pt]
  \item \textbf{Retrieval Coverage:} How often do top-$k$ retrieved passages fully cover the answer span?
  \item \textbf{Precision vs. Recall Tradeoff:} Which chunk size (500 vs. 1000) better balances completeness and retrieval accuracy?
  \item \textbf{Scalability:} What chunk size is more suitable for real workloads?
\end{itemize}

\subsection{Datasets}

\begin{table}[h!]
\centering
\small % <-- 在这里添加字号命令 (其他可选：\footnotesize, \scriptsize)
\caption{LegalBench subsets used in the benchmark}
\label{tab:datasets}
\renewcommand{\arraystretch}{1.2}
\setlength{\tabcolsep}{6pt}
\begin{tabular}{@{}l*{4}{S[table-format=4.0]}@{}}
\toprule
\textbf{Metric} & \textbf{PrivacyQA} & \textbf{CUAD} & \textbf{MAUD} & \textbf{ContractNLI} \\
\midrule
Q--A Pairs & 712 & 1012 & 1215 & 3950 \\
\bottomrule
\end{tabular}
\end{table}
\subsection{Quantitative Results: Chunk Comparison}
The following tables compare Recall and Precision scores for eSapiens using two different chunk sizes (500 and 1000) across all four datasets.

\begin{table*}[t]
\centering
\caption{Performance of eSapiens (Chunk = 500): Recall and Precision at Different $k$}
\label{tab:chunk500}
\renewcommand{\arraystretch}{1.1}
\setlength{\tabcolsep}{4pt}
\resizebox{\textwidth}{!}{
\begin{tabular}{lcccccc|cccccc}
\toprule
\textbf{Dataset} & \multicolumn{6}{c|}{\textbf{Recall@k (\%)}} & \multicolumn{6}{c}{\textbf{Precision@k (\%)}} \\
\cmidrule(lr){2-7} \cmidrule(lr){8-13}
 & k=1 & k=2 & k=4 & k=8 & k=16 & k=50 & k=1 & k=2 & k=4 & k=8 & k=16 & k=50 \\
\midrule
PrivacyQA   & 18.15 & 25.87 & 49.28 & 64.07 & 85.63 & 96.47 & 18.50 & 14.02 & 13.18 & 9.26 & 4.74 & 5.28 \\
ContractNLI & 4.91  & 9.33  & 16.09 & 25.83 & 35.04 & 46.90 & 5.08  & 5.59  & 5.04  & 3.67 & 2.52 & 1.75 \\
MAUD        & 0.52  & 2.48  & 4.39  & 7.24  & 14.03 & 22.60 & 1.94  & 2.63  & 2.05  & 1.77 & 1.79 & 1.12 \\
CUAD        & 3.17  & 7.33  & 18.26 & 28.67 & 42.50 & 55.66 & 3.53  & 4.18  & 6.18  & 5.06 & 3.93 & 2.74 \\
ALL         & 7.26  & 11.52 & 20.40 & 27.94 & 41.37 & 51.82 & 7.49  & 6.82  & 6.65  & 5.02 & 3.78 & 2.68 \\
\bottomrule
\end{tabular}
}
\end{table*}

\begin{table*}[t]
\centering
\caption{Performance of eSapiens (Chunk = 1000): Recall and Precision at Different $k$}
\label{tab:chunk1000}
\renewcommand{\arraystretch}{1.1}
\setlength{\tabcolsep}{4pt}
\resizebox{\textwidth}{!}{
\begin{tabular}{lcccccc|cccccc}
\toprule
\textbf{Dataset} & \multicolumn{6}{c|}{\textbf{Recall@k (\%)}} & \multicolumn{6}{c}{\textbf{Precision@k (\%)}} \\
\cmidrule(lr){2-7} \cmidrule(lr){8-13}
 & k=1 & k=2 & k=4 & k=8 & k=16 & k=50 & k=1 & k=2 & k=4 & k=8 & k=16 & k=50 \\
\midrule
PrivacyQA   & 10.10 & 20.24 & 28.84 & 54.95 & 71.44 & 94.47 & 8.97  & 10.31 & 7.81  & 7.34  & 5.16 & 2.64 \\
ContractNLI & 4.81  & 8.72  & 12.62 & 17.85 & 25.54 & 39.78 & 2.28  & 2.47  & 1.84  & 1.33  & 0.89 & 0.42 \\
MAUD        & 0.52  & 2.48  & 3.05  & 4.57  & 7.31  & 13.60 & 1.33  & 1.07  & 0.84  & 0.64  & 0.53 & 0.32 \\
CUAD        & 3.62  & 10.47 & 20.63 & 32.46 & 45.24 & 62.30 & 2.12  & 3.08  & 3.17  & 2.70  & 2.01 & 0.96 \\
ALL         & 4.93  & 10.34 & 16.29 & 27.46 & 37.38 & 52.54 & 3.68  & 4.23  & 3.42  & 3.00  & 2.15 & 1.09 \\
\bottomrule
\end{tabular}
}
\end{table*}

\subsection{Chunk Size Analysis}

\noindent
\textbf{PrivacyQA:} Due to fragmented original text, chunk = 500 yields higher recall at low $k$, but performance converges with chunk = 1000 as $k$ increases.

\noindent
\textbf{Other datasets:} For \textit{CUAD}, \textit{MAUD}, and \textit{ContractNLI}, chunk = 1000 consistently outperforms 500 in recall, especially at top-50. Larger chunks preserve more semantic information.

\noindent
\textbf{Precision:} While more affected by chunk size, precision is less critical than recall in business applications. Strong models like GPT-4o can filter irrelevant chunks.

\noindent
\textbf{Real-world usage:} Chunk = 1000 is better aligned with production needs, offering broader coverage and lower information loss.

\subsection{Summary}

Through comparative experiments on four datasets — \textit{PrivacyQA}, \textit{CUAD}, \textit{MAUD}, and \textit{ContractNLI} — clear trends are observed in how chunk window size affects retrieval performance.

\noindent\textbf{PrivacyQA:}  
Due to the fragmented nature of its original text, \textit{PrivacyQA} benefits from smaller window sizes. With chunk = 500, vector search achieves higher location precision, leading to superior Recall@k at low $k$ values. However, as $k$ increases, the recall performance of both 500 and 1\,000 converge.

\noindent\textbf{Other datasets:}  
For \textit{CUAD}, \textit{MAUD}, and \textit{ContractNLI}, chunk = 1\,000 consistently outperformed 500 in recall, especially at high $k$ values (e.g., $k=50$). This shows that larger chunks better preserve semantic context and retrieve more useful information.

\noindent\textbf{Precision considerations:}  
Although precision varied more significantly across chunk sizes, it is found that precision is less critical than recall in practical workflows. Given the reasoning capabilities of large models such as GPT‑4o, low-precision context can still be filtered intelligently.

\noindent\textbf{Real-world usage:}  
In enterprise applications where completeness and recall are paramount, chunk = 1\,000 proves more advantageous. While chunk = 500 may slightly improve precision, chunk = 1\,000 delivers a better recall–quality tradeoff.

\section{TRACe-Based Output Quality Evaluation}
\label{sec:appendix-b}

\subsection{Objectives and Setup}
This experiment evaluates the generation quality of eSapiens and a baseline FAISS-based DEREK pipeline using the TRACe framework. Five popular LLMs (GPT-4o, GPT-4o-mini, Claude 3.7, Gemini 1.5 Pro, DeepSeek R1) are tested under identical question sets.

\noindent
Key evaluation dimensions include:
\begin{itemize}[leftmargin=*,itemsep=2pt]
  \item \textbf{Completeness:} Whether key answer points in context were reflected in the output.
  \item \textbf{Utilization:} Proportion of retrieved context actually used.
  \item \textbf{Context Relevance:} Alignment between retrieved content and user intent.
  \item \textbf{pc hallucinated:} Percent of hallucinated tokens.
  \item \textbf{Accuracy:} Human-graded alignment with ground truth.
\end{itemize}

\subsection{Results Overview}
Table~\ref{tab:trace_results} summarizes the TRACe evaluation metrics for each model across the eSapiens and FAISS pipelines.

\begin{table*}[htbp]
\caption{Evaluation results of DEREK models on 100 random questions from RAGtruth}
\centering
\label{tab:trace_results}
\resizebox{\textwidth}{!}{%
\begin{tabular}{lcccccc}
\toprule
\textbf{Model} & \textbf{Completeness} & \textbf{Utilization} & \textbf{Context Relevance} & \textbf{pc hallucinated} & \textbf{Accuracy} \\
\midrule
eSapiens-gpt4o & 0.4307 & 0.5224 & 0.3648 & 0.1823 & 3.85 \\
eSapiens-gpt4o-mini & 0.3433 & 0.4658 & 0.3785 & 0.2729 & 3.70 \\
eSapiens-claude-3.7 & 0.3840 & 0.4440 & 0.3648 & 0.1403 & 4.05 \\
eSapiens-gemini-1.5-pro & 0.4982 & 0.5179 & 0.3728 & 0.1712 & 4.0 \\
eSapiens-deepseek-r1 & 0.4571 & 0.5167 & 0.2581 & 0.1486 & 4.15 \\
faiss+top-2+short+gpt4o & 0.4450 & 0.4800 & 0.3294 & 0.0875 & 3.55 \\
faiss+top-2+short+gpt4o-mini & 0.4105 & 0.4467 & 0.3090 & 0.1524 & 3.15 \\
faiss+top-2+short+claude-3.7 & 0.4985 & 0.5102 & 0.3270 & 0.0860 & 3.75 \\
faiss+top-2+short+gemini-1.5-pro & 0.5346 & 0.5728 & 0.3296 & 0.1371 & 3.75 \\
faiss+top-2+short+deepseek-r1 & 0.5381 & 0.5102 & 0.3430 & 0.1139 & 4.15 \\
\bottomrule
\end{tabular}%
}

\end{table*}

\subsection{Findings and Analysis}
\noindent
\textbf{1. pc hallucinated (hallucination rate)} \\ 
\textit{Findings:} FAISS baseline consistently achieves lower hallucination rates than eSapiens. \\ 
\textit{Explanation:} The baseline uses an explicit instruction to "not answer beyond the given context," prompting models to return "I don't know" when insufficient information is retrieved. The eSapiens platform, while precise, allows final-stage model generation which may hallucinate based on general LLM priors.

\noindent
\textbf{2. Completeness} \\ 
\textit{Findings:} FAISS baseline yields higher completeness scores overall. \\ 
\textit{Explanation:} By constraining generation to retrieved top-$k$ passages, FAISS ensures dense coverage. eSapiens sometimes applies structural formatting or abstraction, leading to partial omission of points.

\noindent
\textbf{3. Context Relevance} \\ 
\textit{Findings:} eSapiens outperforms FAISS in contextual alignment. \\ 
\textit{Explanation:} Even with same $k$, eSapiens uses hybrid retrieval and reranking strategies that produce more targeted, intent-aligned contexts.

\noindent
\textbf{4. Utilization} \\ 
\textit{Findings:} Gemini-1.5-pro and GPT-4o show better context utilization on eSapiens. \\ 
\textit{Explanation:} These models are strong at reasoning over retrieved paragraphs. Gemini performs structured summarization well; GPT-4o excels in textual citation.

\noindent
\textbf{5. Accuracy (human judgment)} \\ 
\textit{Findings:} eSapiens wins on user-rated naturalness and coherence. \\ 
\textit{Explanation:} eSapiens integrates formatting and rephrasing post-processing, improving readability while preserving core meaning.

\subsection{Summary}
In summary, eSapiens shows clear gains in relevance, fluency, and model alignment—particularly when paired with Gemini or GPT-4o. The FAISS baseline yields stricter factual alignment with fewer hallucinations but lacks the naturalness and flexibility found in eSapiens generations. Each pipeline offers tradeoffs that may suit different use cases (e.g., regulatory Q\&A vs. creative summarization).

\section{Natural-Language Prompts \& Full SQL Listings}
\label{app:sql}

\subsection*{Prompt 1}
\textbf{Natural-language question}\\
\emph{“Show me the pending deliveries by month for the last 18 months.”}

\textbf{SQL queries generated by Product X}
\begin{lstlisting}
SELECT
  DATE_TRUNC('month', delivery_requests.created_at) AS month,
  COUNT(delivery_requests.id)                       AS pending_deliveries
FROM payload_catalog.public.delivery_requests
WHERE delivery_requests.status ILIKE '%pending%'
  AND delivery_requests.created_at >= DATE_SUB(CURRENT_DATE(), 18 * 30)
GROUP BY month
ORDER BY month ASC;
\end{lstlisting}

\textbf{SQL queries generated by eSapiens' THOR}
\begin{lstlisting}
WITH monthly_deliveries AS (
  SELECT DATE_TRUNC('month', created_at) AS month,
         status,
         COUNT(*)                        AS delivery_count
  FROM   delivery_requests
  WHERE  created_at >= NOW() - INTERVAL '18 months'
    AND  created_at <= NOW()
    AND  status IS NOT NULL
  GROUP  BY DATE_TRUNC('month', created_at), status
)
SELECT TO_CHAR(month,'YYYY-MM') AS month,
       status,
       delivery_count
FROM   monthly_deliveries
ORDER  BY month DESC, status;
\end{lstlisting}

%-------------------------------------------------------
\subsection*{Prompt 2}
\textbf{Natural-language question}\\
\emph{“What are the different channels through which drivers learned about Payeload, and what is the percentage of drivers for each channel? In addition, analyze these data to provide actionable insights and recommendations for the marketing team to optimize their outreach strategy. '}

\textbf{SQL queries generated by Product X}
\begin{lstlisting}
SELECT
  users.invitation_token,
  ( COUNT(users.id) * 100.0 /
    (SELECT COUNT(*) FROM payload_catalog.public.users)
  ) AS percentage
FROM payload_catalog.public.users
GROUP BY users.invitation_token;
\end{lstlisting}

\textbf{SQL queries generated by eSapiens' THOR}
\begin{lstlisting}
WITH channel_counts AS (
  SELECT
    COALESCE(how_did_you_hear,'Not Specified') AS channel,
    COUNT(*)                                     AS driver_count,
    COUNT(*) * 100.0 /
      (SELECT COUNT(*) FROM driver_details)   AS percentage
  FROM driver_details
  GROUP BY how_did_you_hear
  ORDER BY driver_count DESC
)
SELECT channel,
       driver_count,
       ROUND(percentage,2) AS percentage
FROM   channel_counts;
\end{lstlisting}

%-------------------------------------------------------
\subsection*{Prompt 3}
\textbf{Natural-language question}\\
\emph{“I want to see the top 10 regions with higher income \textbf{per mile} by region in the past 3 months. Note: The distance field in the \texttt{delivery\_requests} table is measured in meters. '}

\textbf{SQL queries generated by Product Y}
\begin{lstlisting}
WITH recent_requests AS (
  SELECT
    dr.id,
    a.region_id,
    dr.fee_total_calculated,
    dr.distance
  FROM payload_catalog.public.delivery_requests dr
  JOIN payload_catalog.public.accounts          a
       ON dr.account_id = a.id
  WHERE dr.created_at >= DATE_SUB(CURRENT_DATE(), 90)
),
region_income AS (
  SELECT
    region_id,
    SUM(fee_total_calculated) /
    SUM(distance)               AS income_per_mile
  FROM recent_requests
  GROUP BY region_id
)
SELECT *
FROM   region_income
ORDER  BY income_per_mile DESC
LIMIT 10;
\end{lstlisting}

\textbf{SQL queries generated by eSapiens' THOR}
\begin{lstlisting}
WITH region_metrics AS (
  SELECT
    r.name                                       AS region_name,
    SUM(dr.fee_total_calculated/1000)            AS total_revenue_dollars,
    SUM(dr.distance / 1609.34)                   AS total_distance_miles,
    CASE
      WHEN SUM(dr.distance / 1609.34) = 0
      THEN 0
      ELSE SUM(dr.fee_total_calculated/1000) /
           SUM(dr.distance / 1609.34)
    END                                          AS revenue_per_mile
  FROM delivery_requests dr
  JOIN accounts          a ON dr.account_id = a.id
  JOIN regions           r ON a.region_id   = r.id
  WHERE dr.created_at >= CURRENT_DATE - INTERVAL '3 months'
    AND dr.status IN ('delivered','completed','DELIVERED','COMPLETED')
  GROUP BY r.name
  HAVING SUM(dr.distance) > 0
)
SELECT region_name,
       ROUND(total_revenue_dollars,2),
       ROUND(total_distance_miles,2),
       ROUND(revenue_per_mile,2) AS dollars_per_mile
FROM   region_metrics
ORDER  BY revenue_per_mile DESC
LIMIT 10;
\end{lstlisting}

%-------------------------------------------------------
\subsection*{Prompt 4}
\textbf{Natural-language question}\\
\emph{“Which track has the highest unit price?”}

\textbf{SQL queries generated by Product Y}
\begin{lstlisting}
SELECT name, unit_price
FROM   chinook_track
ORDER  BY unit_price DESC
LIMIT 1;
\end{lstlisting}

\textbf{SQL queries generated by eSapiens' THOR}
\begin{lstlisting}
-- identical query; eSapiens adds a narrative explanation
SELECT name, unit_price
FROM   chinook_track
ORDER  BY unit_price DESC
LIMIT 1;
\end{lstlisting}

%-------------------------------------------------------
\subsection*{Prompt 5}
\textbf{Natural-language question}\\
\emph{“How many hip-hop tracks are there in the database?” (requires fuzzy matching).}

\textbf{SQL queries generated by Product Z}
\begin{lstlisting}
SELECT COUNT(*) AS hip_hop_tracks
FROM   chinook_track
WHERE  LOWER(genre) = 'hip hop';
\end{lstlisting}

\textbf{SQL queries generated by eSapiens' THOR}
\begin{lstlisting}
SELECT COUNT(*) AS hip_hop_track_count
FROM   chinook_track
WHERE  LOWER(genre) LIKE '%hiphop%'
   OR  LOWER(genre) LIKE '%hip hop%'
   OR  LOWER(genre) LIKE '%hip-hop%';
\end{lstlisting}

%-------------------------------------------------------
\subsection*{Prompt 6}
\textbf{Natural-language question}\\
\emph{“Provide the sales data for the past three months.”
Note: There is future sales data in the dataset.}

\textbf{SQL queries generated by Product Z}
\begin{lstlisting}
SELECT DATE_TRUNC('month', invoice_date) AS month,
       SUM(total)                        AS total_sales
FROM   chinook_invoice
WHERE  invoice_date >= (CURRENT_TIMESTAMP - INTERVAL '3 months')
GROUP  BY month
ORDER  BY month DESC;
\end{lstlisting}

\textbf{SQL queries generated by Product Z}
\begin{lstlisting}
WITH __chinook_invoice AS (
  SELECT DATE_TRUNC('MONTH', invoice_date) AS month,
         SUM(total)                        AS total_sales
  FROM   test.chinook.chinook_invoice
  WHERE  invoice_date >= DATEADD(MONTH, -3, CURRENT_DATE)
  GROUP  BY DATE_TRUNC('MONTH', invoice_date)
)
SELECT *
FROM   __chinook_invoice
ORDER  BY month DESC;
\end{lstlisting}
\textbf{SQL queries generated by eSapiens' THOR}
\begin{lstlisting}
SELECT
  i.invoice_id,
  i.customer_id,
  i.invoice_date,
  i.total,
  c.first_name || ' ' || c.last_name AS customer_name,
  il.track_id,
  t.name                             AS track_name,
  il.unit_price,
  il.quantity,
  (il.unit_price * il.quantity)      AS line_total
FROM  chinook_invoice      i
JOIN  chinook_customer     c  ON i.customer_id = c.customer_id
JOIN  chinook_invoice_line il ON i.invoice_id  = il.invoice_id
LEFT JOIN chinook_track    t  ON il.track_id   = t.track_id
WHERE i.invoice_date BETWEEN '2025-01-17' AND '2025-04-17'
ORDER BY i.invoice_date DESC, i.invoice_id;
\end{lstlisting}

%-------------------------------------------------------

\end{document}